\documentclass[11pt,a4paper]{article}

\usepackage{authblk}
\usepackage[hyperref]{acl2020}
\usepackage{times}
\usepackage{latexsym}

\usepackage{microtype}
\usepackage{amsmath,amssymb,amsfonts}
\usepackage{caption}
\usepackage{algorithmic}
\usepackage{array}
\usepackage{graphicx}
\usepackage{textcomp}
\usepackage{xcolor}
\usepackage{url}
\usepackage{siunitx}
\usepackage{pgfplots}
\pgfplotsset{compat=newest}
\usepackage{booktabs}
\usepackage{multirow}
\usepackage{float}
\usepackage{relsize}
\floatstyle{plaintop}
\restylefloat{table}

\aclfinalcopy 
\def\aclpaperid{13} 

\title{EmpLite: A Lightweight Sequence Labeling Model for Emphasis Selection of Short Texts}

\author{Vibhav Agarwal, Sourav Ghosh, Kranti Chalamalasetti, Bharath Challa,\\Sonal Kumari, Harshavardhana, Barath Raj Kandur Raja}
\affil{\{vibhav.a, sourav.ghosh, kranti.ch, bharath.c, sonal.kumari,\\ harsha.vp, barathraj.kr\}@samsung.com}
\affil{Samsung R\&D Institute Bangalore, Karnataka, India 560037}

\date{}

\begin{document}
\maketitle
\begin{abstract}
	
	Word emphasis in textual content aims at conveying the desired intention by changing the size, color, typeface, style (bold, italic, etc.), and other typographical features. The emphasized words are extremely helpful in drawing the readers' attention to specific information that the authors wish to emphasize. However, performing such emphasis using a soft keyboard for social media interactions is time-consuming and has an associated learning curve. In this paper, we propose a novel approach to automate the emphasis word detection on short written texts. To the best of our knowledge, this work presents the first lightweight deep learning approach for smartphone deployment of emphasis selection. Experimental results show that our approach achieves comparable accuracy at a much lower model size than existing models. Our best lightweight model has a memory footprint of 2.82 MB with a matching score of 0.716 on SemEval-2020 \cite{semeval2020-task10} public benchmark dataset.
	
\end{abstract}


\section{Introduction}\label{sec:introduction}

\begin{figure}
		\centering
		\includegraphics[width=0.75\linewidth]{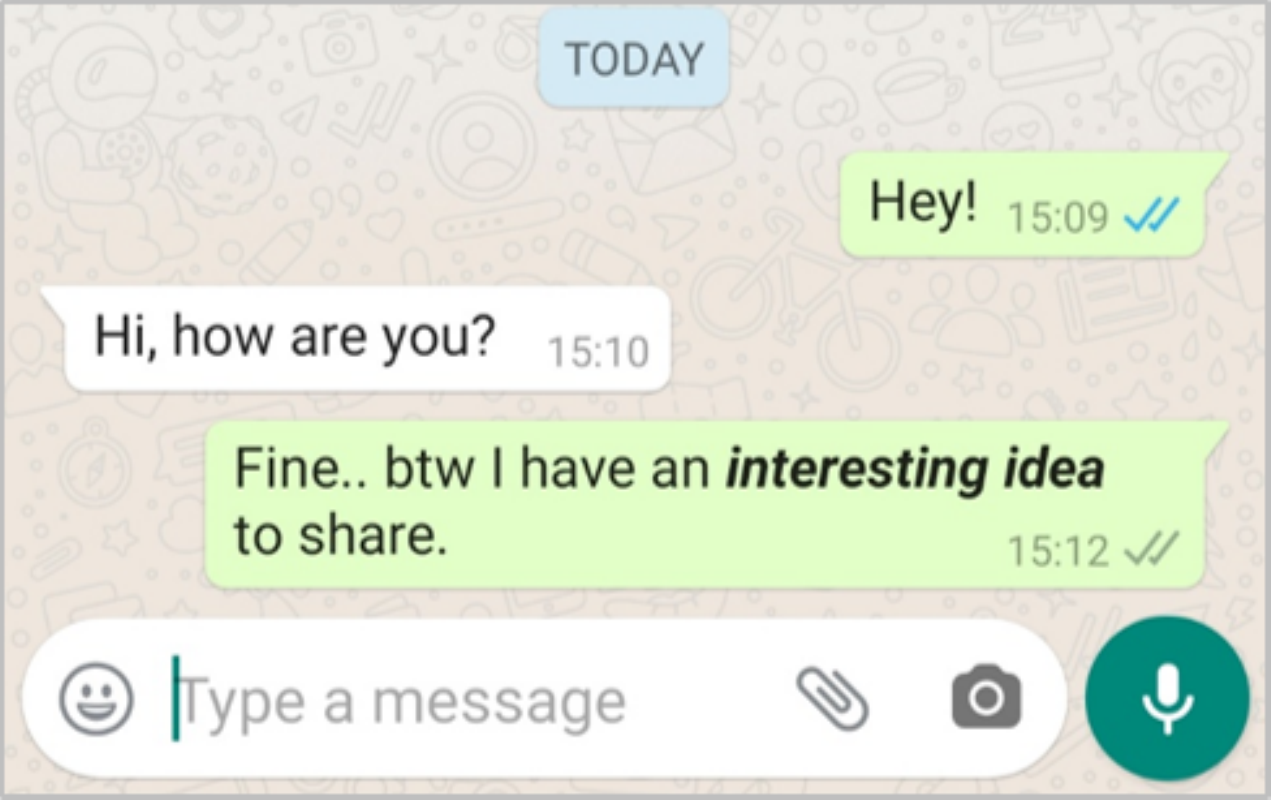}
		\caption{Prominent words in a message are being emphasized (\textbf{Bold} + \textit{Italic}) automatically}
		\label{fig:emphasis_message}
	\end{figure}

	Emphasizing words or phrases is commonly performed to drive a point strongly and/or to highlight the key terms and phrases. While speaking, speakers can use tone, pitch, pause, repetition, etc. to highlight the core of a speech in the minds of an audience. Similarly, while writing or messaging, authors can emphasize the words by customizing the formatting like typeface, font size, bold, italic, font color, etc. (Figure \ref{fig:emphasis_message}). With the explosion of social media, word emphasis has become more critical in engaging readers' attention and conveying the author's message in the shortest possible time.
	
	Emphasis selection of text has recently emerged as a focus of research interest in natural language processing (NLP). The goal of emphasis selection is to automate the identification of words or phrases that bring clarity and convey the desired meaning. Automatic emphasis selection can help in better graphic designing and presentation applications, as well as can enable voice assistants and digital avatars to realize expressive text-to-speech (TTS) synthesis. High-quality emphasis selection models can enable automatic design assistance for creating flyers, posters and accelerate the workflow of design programs such as Adobe Spark \cite{adobe-spark}, Microsoft PowerPoint, etc. These emphasis selection models can also empower digital avatars like Samsung Neon \cite{samsung-neon} to achieve human-like TTS systems. Understanding emphasis selection is also crucial for many downstream applications in NLP tasks including text summarization, text categorization, information retrieval, and opinion mining.
	\looseness=-1
	
	In the current work, we propose a novel lightweight neural architecture for automatic emphasis selection in short texts, which can perform inference in a low-resource constrained environment like a smartphone. Our proposed architecture achieves near-SOTA performance, with as low as 0.6\% of its model size.
	\looseness=-1

\section{Related Work}\label{sec:relatedWork}
	
	Prior work in NLP literature towards identifying important words or phrases have focused widely on \textit{keyword or key-phrase extraction}. Considerable progress has been made in keyword or key-phrase extraction systems for long documents such as news articles, scientific publications, etc. \cite{rose2010automatic}. The core operation procedure of these systems is to extract the nouns and noun phrases. To achieve these, researchers have used statistical co-occurrence \cite{matsuo2004keyword}, SVM \cite{zhang2006keyword}, CRF \cite{zhang2008automatic}, graph-based extraction \cite{litvak2008graph}, etc. Recent efforts have even expanded the idea from a set of documents to social big data \cite{kim2020efficient}. However, in the context of short texts like text messages, headlines, or quotes, keyword extraction systems often mislabel most nouns as important without considering the essence of the text, thus performing poorly at the task.

	\textit{Emphasis selection} aims to overcome this by scoring words which properly capture the essence of a text by focusing on subtle cues of emotions, clarifications, and words that capture readers' attention, as seen in Table \ref{tab:keywordExtractionVsEmphasisSelection}. Recent research interest towards these tasks often uses label distribution learning \cite{shirani2019learning}. MIDAS \cite{anand2020midas} uses label distribution as well as contextual embeddings. One drawback of using label distribution learning is the requirement of annotations, which are not readily available in most datasets. Pre-trained language model has also been used to achieve emphasis selection \cite{huang2020ernie}. Singhal et al. \cite{singhal2020iitk} achieves significantly good performance with (a) Bi-LSTM + Attention approach, and (b) Transformers approach. To achieve their modest performances, these architectures produce huge models. For instance, IITK model \cite{singhal2020iitk} takes up 469.20 MB in BiLSTM + Attention approach, while requiring almost 1.5 GB in Transformers approach. This is partly due to the use of embeddings like BERT (1.2 GB) \cite{devlin2018bert}, XLNET (1.34 GB) \cite{yang2019xlnet}, RoBERTa (1.3 GB) \cite{liu2019roberta}, etc. General-purpose models that emphasize on model size still consume significant ROM: 200 MB (for DistilBERT \cite{sanh2019distilbert}) and 119 MB (for MobileBERT \cite{sun2020mobilebert} quantized int8 saved model and variables; sequence length 384). Thus, in spite of the performance benefits, these emphasis selection systems with high-memory footprints are not suitable for the storage specifications of mobile devices.
		
	Thus, while keyword extraction systems are not suitable for short text content, emphasis selection systems perform much better at such tasks. However, most existing architectures of the latter are not light-weight, and thus, not suitable for on-device inferencing on low-resource devices. This motivates us to propose EmpLite, which (a) outperforms keyword extraction systems by using emphasis selection for use with short texts, and (b) differs from existing emphasis selection architectures by ensuring a very light-weight model for efficient on-device inferencing on mobile devices. Our decisions towards achieving low model size include using a subset of GloVe \cite{pennington-etal-2014-glove} word embeddings, thus, reducing embedding size from 347.1 MB to 2.5 MB, which we discuss in section \ref{subsec:characterAndWordLevelFeatures}.
	
	\begin{table}[t]	
		\small
		\centering
		\resizebox{\columnwidth}{!}{%
			\begin{tabular}{p{3cm}|p{3cm}|p{3cm}}
				\toprule
				\textbf{Input Text} & \textbf{Keywords/Key phrases Detected} & \textbf{Emphasis Selection} \\ \midrule
				
				A simple I love you means more than money & A simple I love you means more than \textbf{money} & A simple \textbf{I love you} means more than money\\ \midrule
				Traveling – It leaves you speechless then turns you into story teller & Traveling – It leaves you \textbf{speechless} then turns you into \textbf{story teller} & \textbf{Traveling} – It leaves you \textbf{speechless} then turns you into \textbf{story teller} \\ \midrule
				Challenges are what make life more interesting and overcoming them is what makes life meaningful & \textbf{Challenges} are what make life more \textbf{interesting} and overcoming them is what makes life \textbf{meaningful} & \textbf{Challenges} are what make life more \textbf{interesting} and \textbf{overcoming} them is what makes life \textbf{meaningful} \\ \bottomrule
			\end{tabular}%
		}
		\caption{Keyword Extraction \cite{monkeylearn-keyword-extraction} vs. Emphasis Selection}\label{tab:keywordExtractionVsEmphasisSelection}
	\end{table}
	
	\begin{table*}[t]	
		\small
		\centering
		\resizebox{2\columnwidth}{!}{%
			\begin{tabular}{c c c c c c c c c c c c}
				\toprule
				\textbf{Word}   & \textbf{A1} & \textbf{A2} & \textbf{A3} & \textbf{A4} & \textbf{A5} & \textbf{A6} & \textbf{A7} & \textbf{A8} & \textbf{A9} & \textbf{Freq [B|I|O]} & \textbf{Emphasis Prob (B+I)/(B+I+O)} \\ \midrule
				Kindness & B  & B  & B  & O  & O  & O  & B  & B  & B  & 6|0|3        & 0.666                       \\
				is    & O  & O  & O  & O  & O  & O  & I  & I  & O  & 0|2|7        & 0.222                       \\
				like   & O  & O  & O  & O  & O  & O  & I  & I  & O  & 0|2|7        & 0.222                       \\
				snow   & O  & O  & B  & O  & O  & O  & I  & I  & O  & 1|2|6        & 0.333                       \\ \bottomrule
			\end{tabular}%
		}
		\caption{A short text example from dataset along with its nine annotations}\label{tab:datasetExampleWithAnnotations}
	\end{table*}

\section{Data Collection}\label{sec:dataCollection}

	We use the officially released SemEval-2020 dataset \cite{SemEval2020-Task10-Emphasis-Selection}, which is the combination of Spark dataset \cite{adobe-spark} and Wisdom Quotes dataset \cite{wisdom-quotes}. The dataset consists of 3,134 samples labeled for token-level emphasis by multiple annotators. There are 7,550 tokens with fewer than 10 words in a sample and they are randomly divided into training (70\%), development (10\%), and test (20\%) sets by the organizers. Table \ref{tab:datasetExampleWithAnnotations} shows a short text example from the training set, annotated with the BIO annotations, where `B (beginning) / I (inside)' and `O (outside)' represent emphasis and non-emphasis words, respectively, as decided by an annotator. The last column shows the emphasis probability for a word, computed as (B+I) divided by the total number of annotators, i.e. 9. We generate data labels for model training using emphasis probabilities by assigning 0 to low emphasis words (having $\text{probability} < \text{threshold}_{prob}$) and 1 to high emphasis words ($\text{probability} \ge \text{threshold}_{prob}$). We experiment with different values for $\text{threshold}_{prob}$ and observe that 0.4 yields the best results.
		
	\subsection{Evaluation Metric}\label{subsec:evaluationMetric}
	
		The evaluation metric for our problem is defined as follows:
		
		\textbf{$\text{Match}_m$} \cite{semeval2020-task10}: For each instance $x$ in the test set $D_{\text{test}}$, we select a set $S^{(x)}_m$ of $m \in (1..4)$ words with the top $m$ probabilities with high emphasis according to the ground truth. Analogously, we select a prediction set $\hat{S}^{(x)}_m$ for each $m$, on the basis of the predicted probabilities. We define matching score, or $\text{Match}_m$, as:
				
		\begin{equation}
			\text{Match}_m = \frac{\sum\limits_{x \in D_{\text{test}}} \left|S^{(x)}_m \cap \hat{S}^{(x)}_m\right| \Big/ m }{\left|D_{\text{test}}\right|}
		\end{equation}
	
	
		Then, we compute the average rank score by averaging all possible $\text{Match}_m$ scores:
		
		\begin{equation}
			\text{Average Score} = \frac{\sum\limits_{m \in (1..4)} \text{Match}_m }{4}
		\end{equation}
	
	\subsection{Data Augmentation}\label{subsec:dataAugmentation}
		There are only 3,134 annotated samples in the training data, which makes it difficult to improve the accuracy with our neural model. So, to enlarge the amount of training data, we experiment with four data augmentation strategies \cite{sun2020ernie}:
		
		\begin{enumerate}
			\item Randomly removing $\le$ 1 word per sentence,
			\item Randomly removing $\ge$ 1 word per sentence,
			\item Upper-casing a word randomly, and
			\item Reversing the sentence.
		\end{enumerate}
		
		The effect of these techniques on our accuracy metric is presented in Section \ref{subsec:dataAugmentationAnalysisResults}.
		
	\section{System Overview}\label{sec:systemOverview}
\begin{figure}
			\centering
			\includegraphics[width=\linewidth]{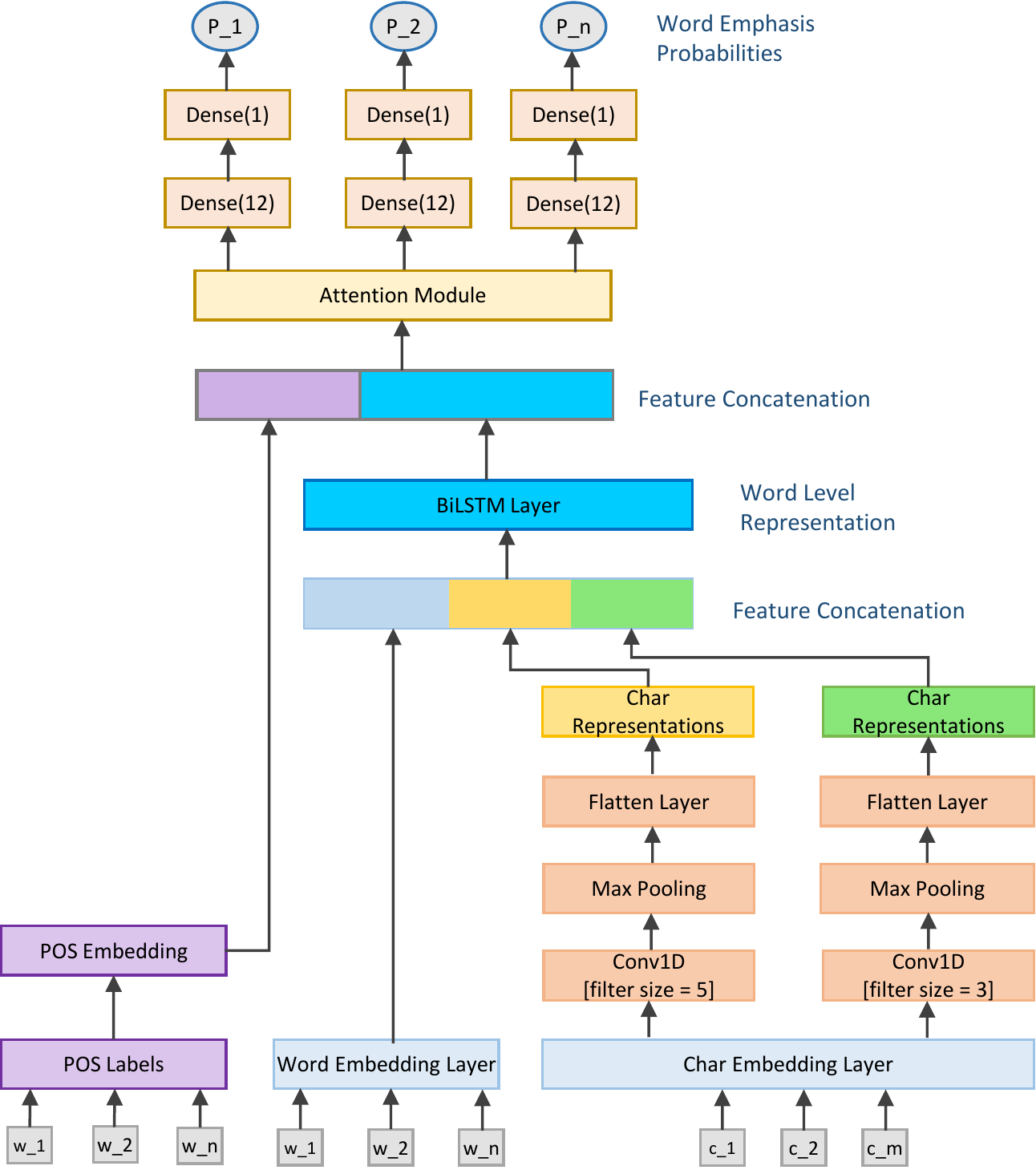}
			\caption{The proposed EmpLite Model Architecture}
			\label{fig:Architecure_diag}
		\end{figure}

		We begin with a basic model and enhance that model with contextual information (in the form of pre-trained embeddings, char-level embeddings, Parts of Speech Tag concatenation, etc.). We describe the key components (layers) of our final EmpLite neural network architecture, as illustrated in Figure \ref{fig:Architecure_diag}.
		
		\subsection{Character and Word level features}\label{subsec:characterAndWordLevelFeatures}
			A combination of word-level and character-level input representations has shown great success for several NLP tasks \cite{liang2017combining}. This is because word representation is suitable for relation classification, but it does not perform well on short, informal, conversational texts, whereas char representation handles such informal texts very well. To take the best of both representations, our proposed EmpLite model employs a combination of word and character encodings for a robust understanding of context.
			
			We used two layers of CNN \cite{ma2016end} with 1D convolutional layers of filter sizes 3 and 5 that extracts multiple character-level representations and handles misspelled words as well as models sub-word structures such as prefixes and suffixes. We use the same number of filters for both convolutional layers: 16, selected on the basis of experimental results for optimal accuracy.
			
			The character level embeddings obtained are then concatenated with pre-trained GloVe \cite{pennington-etal-2014-glove} 50-dimensional word embeddings. Here, we use a subset of the GloVe embeddings corresponding to training set vocabulary (4331 words), bringing the embedding size down to 2.5 MB. We do not use ELMo \cite{peters2018deep} or other deep contextual embeddings, as it is not feasible to port these heavy pre-trained models for on-device inferencing. In order to handle words not part of training vocabulary, we use a representation, \texttt{<UNK>} token. We set the word embedding layer as trainable as that yields the best score due to fine-tuning of layer weights on our task. The $i^{\text{th}}$ word encoding, $o_{w_i}$, is computed as:
			\looseness=-1
			
			\begin{multline}
				o_{w_i} = \operatorname{concat} \Big(e_{w_i}, \text{CNN}_1 \left(e_{c_1}, e_{c_2}, ..., e_{c_n}\right), \\ \text{CNN}_2 \left(e_{c_1}, e_{c_2}, ..., e_{c_n}\right)\Big)
			\end{multline}
		
			where, $e_{w_i}$ is the word embedding for each word, $w_i$, in the dataset and $e_{c_i}$ is the character embedding for the input character $c_i$.
			
			\begin{table*}[t]	
				\small
				\centering
				\resizebox{2\columnwidth}{!}{%
					\begin{tabular}{m{0.8\columnwidth} c c c c c c}
						\toprule
						\multirow{2}{*}{\textbf{Model}}                                                                                                       & \multirow{2}{*}{\textbf{Model Size (MB)}} &               \multicolumn{4}{c}{\textbf{$\text{Match}_m$}}               & \multirow{2}{*}{\textbf{Average Score}} \\
						\cmidrule{3-6}                                                                                                                        &                                           & $\mathbf{m=1}$   & $\mathbf{m=2}$   & $\mathbf{m=3}$   & $\mathbf{m=4}$   &                                         \\ \midrule
						\textit{Base:} Word\_Emb + BiLSTM + Dense Layer                                                                                       & 1.10                                      & 0.479            & 0.639            & 0.731            & 0.785            & 0.659                                   \\
						[2ex] \cmidrule{1-1}
						Concat[\textbf{LSTM(Char\_Emb)} + Word\_Emb] + BiLSTM + Dense Layer                                         & 1.10                                      & 0.473            & 0.658            & 0.739            & 0.786            & 0.664                                   \\
						[2ex] \cmidrule{1-1}
						Concat[LSTM(Char\_Emb) + \textbf{Word\_GloVe} (Non-trainable)] + BiLSTM + Dense Layer                       & 1.02                                      & 0.514            & 0.660            & 0.748            & 0.795            & 0.679                                   \\
						[2ex] \cmidrule{1-1}
						Concat[\textbf{CNN}(Char\_Emb) + Word\_GloVe (Non-trainable)] + BiLSTM + Dense Layer                        & 1.04                                      & 0.523            & 0.669            & 0.754            & 0.801            & 0.687                                   \\
						[2ex] \cmidrule{1-1}
						Concat[CNN(Char\_Emb) + Word\_GloVe (\textbf{Trainable})] + BiLSTM + Dense Layer                            & 2.70                                      & 0.538            & 0.680            & 0.766            & 0.811            & 0.699                                   \\
						[2ex] \cmidrule{1-1}
						Concat[CNN$_1$ (Char\_Emb) + \textbf{CNN$_2$ (Char\_Emb)} + Word\_GloVe (Trainable)] + BiLSTM + Dense Layer & 2.77                                      & 0.528            & 0.690            & 0.771            & 0.810            & 0.701                                   \\
						[2ex] \cmidrule{1-1}
						Above Model + \textbf{Attention}                                                                            & 2.80                                      & $\mathbf{0.549}$ & 0.684            & 0.779            & 0.817            & 0.707                                   \\
						[2ex] \cmidrule{1-1}
						\textit{EmpLite:} Above Model + \textbf{POS Feature} Concatenation                                          & 2.82                                      & 0.541            & $\mathbf{0.698}$ & $\mathbf{0.782}$ & $\mathbf{0.823}$ & $\mathbf{0.711 }$                       \\ \bottomrule
					\end{tabular}%
				}
				\caption{Comparison of different model architectures}\label{tab:comparisonOfModelArchitectures}
			\end{table*}
			
		\subsection{Word level BiLSTM}\label{subsec:wordLevelBiLSTM}
			The concatenated word representations obtained are then passed through a BiLSTM \cite{10.1162/neco.1997.9.8.1735} layer with 16 units. The BiLSTM layer extracts the features from both forward and backward directions and concatenates the output vectors from each direction. Also, regular and recurrent dropouts with value 0.2 are applied to reduce model overfitting. Let $\overrightarrow{r}$ and $\overleftarrow{r}$ be the forward and backward output states of the BiLSTM. Then, the output vector, $r_b$, is defined as:
			
			\begin{equation}
				r_b = \overrightarrow{r} \oplus \overleftarrow{r}
			\end{equation}
		
			\begin{figure}
				\centering
				\includegraphics[width=\linewidth]{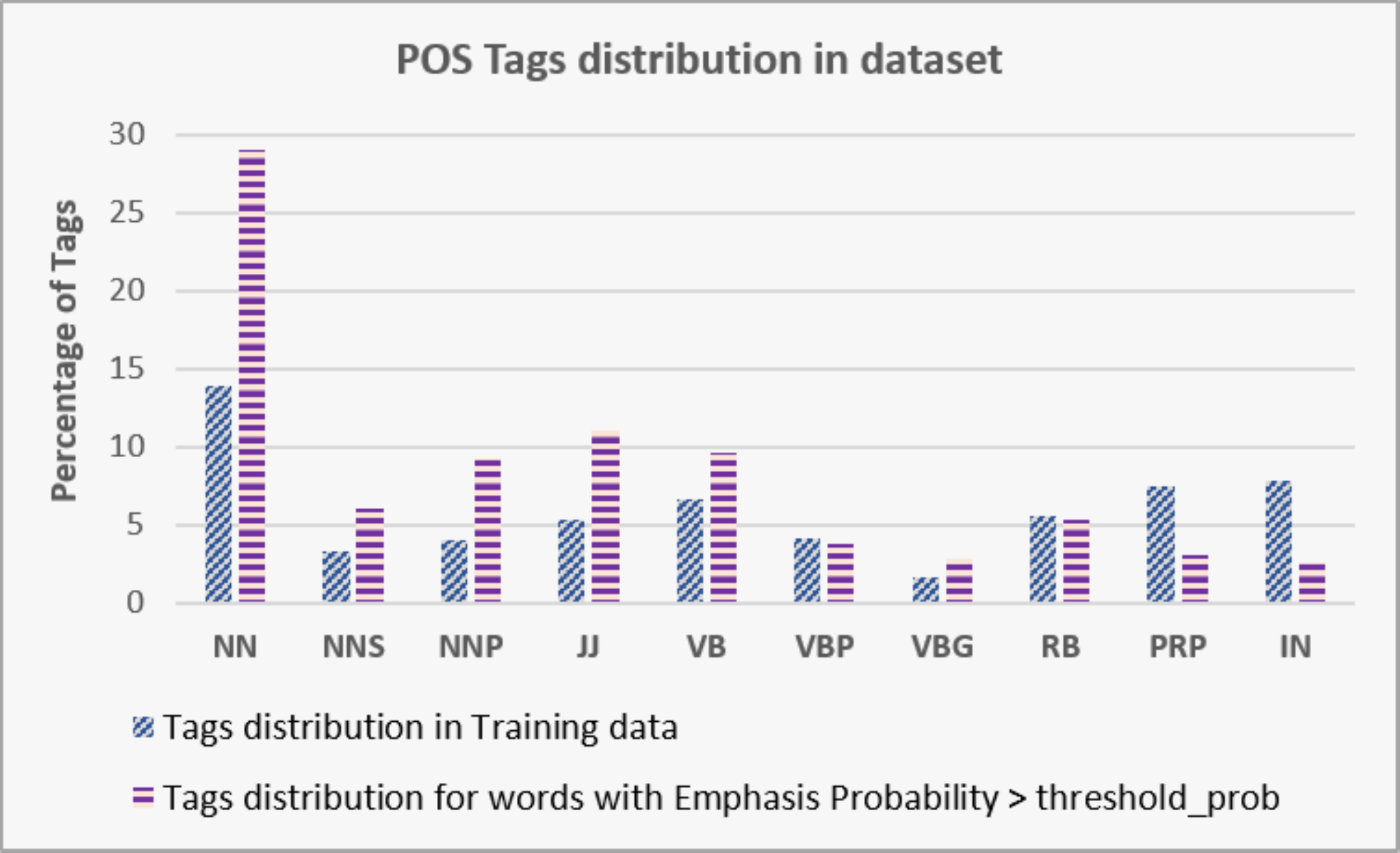}
				\caption{Percentage distribution of top POS tags in training data and for words with emphasis probability greater than threshold}
				\label{fig:POS_distribution_Plot}
			\end{figure}
		
		\subsection{Part of Speech (POS) Tag feature}\label{subsec:posTagFeature}
			Figure \ref{fig:POS_distribution_Plot} illustrates occurrence of top 10 POS Tags \cite{10.3115/1075812.1075835} in our training data. We can infer that POS acts as an important input modeling feature as words with POS Tag: Noun (NN, NNP, NNS), Adjective (JJ) or Verb (VB, VBP, VBG) usually have high emphasis probability whereas Pronouns (PRP) and Prepositions (IN) are less likely to be emphasized. Therefore, we use 16-dimensional embedding to encode POS tag information, which is concatenated with the output of the Bi-LSTM layer (obtained from Equation 4):
			
			\begin{equation}
				\overrightarrow{h} = \operatorname{concat} \left( r_b, e_{pos} \right)
			\end{equation}
		
			where, $e_{pos}$ is the POS feature embedding for the sequence.

		\subsection{Attention Layer}\label{subsec:attentionLayer}
			We add the attention \cite{vaswani2017attention} layer to effectively capture prominent words in the input text sequence. The attention weight is computed as the weighted sum of the output of the previous layer, as shown below:
			
			\begin{equation}
				Z = \operatorname{softmax} \left(w^T \left( \tanh\left({\overrightarrow{h_1}, \overrightarrow{h_2}, ..., \overrightarrow{h_i}, ..., \overrightarrow{h_n}}\right)\right)\right)
			\end{equation}
		
			where, $\overrightarrow{h_i}$ represents output vector of the previous layer, and $w^T$ is the transpose of the trained parameter vector.
			
			The attention layer output is then passed through two time-distributed dense layers with 12 and 1 units, respectively, with $\operatorname{sigmoid}$ activation function to output emphasis probability with respect to each word.

	\section{Experimental Settings \& Results}\label{sec:experimentalSettingsAndResults}
		We attempt numerous small changes to our model to enhance the performance. We choose the hyperparameters to optimize accuracy while maintaining a small model size. As the proposed solution is for mobile devices, we have also captured a system-specific metric, the model size in MB. We use the Tensorflow framework \cite{abadi2016tensorflow} for building the models. Table \ref{tab:comparisonOfModelArchitectures} shows the comparison of $\text{Match}_m$ scores across different variants of lightweight models evaluated on test data.
		
		\vfill
		
		The total number of trainable parameters vary in the range of 21,574 to 238,620 for all the model results reported in Table 3. We train the models with 32 batch-size and compile the model using Adam optimizer \cite{kingma2014adam}. We observe that using CNN gives a better score as compared to LSTM because varying the size of kernels (3 and 5) and concatenating their outputs allow the model to detect patterns of multiple sizes.
		
		\vfill

		We can also infer that using 50-dimensional GloVe embeddings and setting it as trainable improves the overall matching score. This is because we are utilizing language semantic knowledge acquired from the pre-trained embeddings and then fine-tuning it for our task. However, there is an increase in model size due to more number of trainable parameters. Furthermore, we observe marginal gains in the matching score by adding POS tag as a feature followed by an attention layer as these help in a better identification of prominent keywords based on the context.

		\begin{table}[h]	
			\small
			\centering
			\resizebox{\columnwidth}{!}{%
				\begin{tabular}{m{4.35cm} S[table-format=1.3] r}
					\toprule
					\textbf{Model}           & \textbf{$\text{Match}_m$} & \textbf{Size (MB)} \\ \midrule
					IITK: BiLSTM + Attention Approach & 0.747                     & 469.20                   \\ [3ex]
					IITK: Transformers Approach    & $\mathbf{0.804}$          & 1536.00                  \\ [3ex]
					EmpLite              & 0.716                     & $\mathbf{2.82}$          \\ \bottomrule
				\end{tabular}%
			}
			\caption{Comparison with SOTA \cite{singhal2020iitk}}\label{tab:comparisonWithSOTA}
		\end{table}
		
		
		\begin{figure*}
			\centering
			\includegraphics[width=0.8\linewidth]{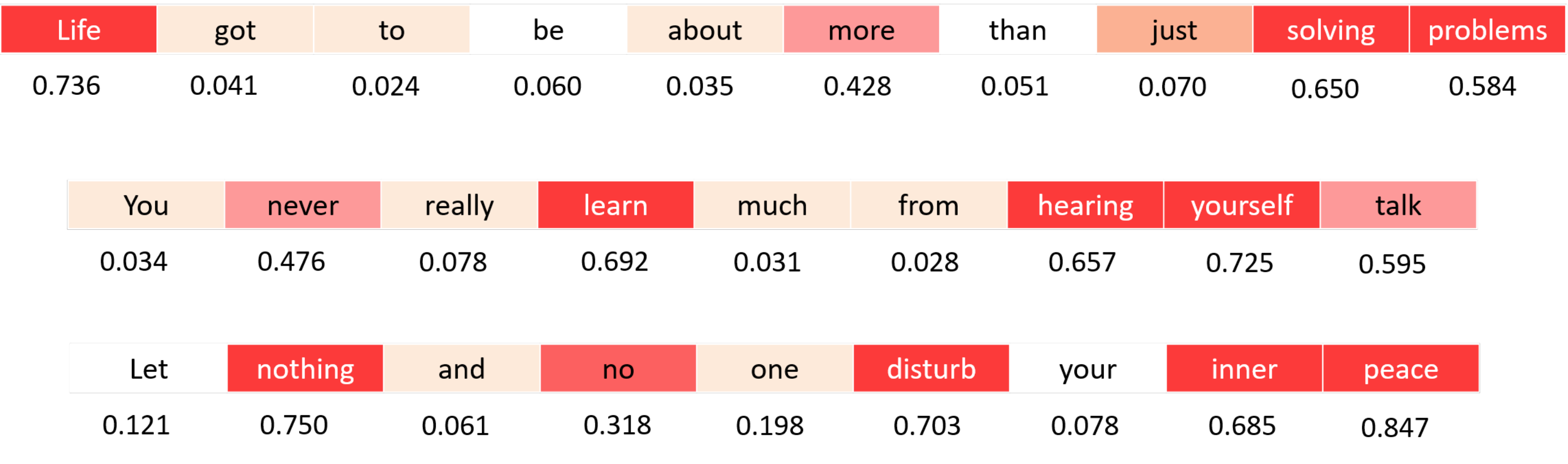}
			\caption{Emphasis Heatmap with word probabilities from EmpLite}
			\label{fig:emphasis_heatmap}
		\end{figure*}
		
		Figure \ref{fig:emphasis_heatmap} presents the Emphasis Heatmap for some examples from the test set using our final model. In Table \ref{tab:comparisonWithSOTA} we benchmark our EmpLite model with the state-of-the-art solution by IITK \cite{singhal2020iitk} which utilized huge pre-trained models like ELMo, BERT \cite{devlin2018bert}, RoBERTa \cite{liu2019roberta} and XLNet \cite{yang2019xlnet}. These models require huge RAM/ROM for on-device inferencing making it unsuitable for edge devices where resources are constrained.

		\begin{table}[h]	
			\small
			\centering
			\resizebox{\columnwidth}{!}{%
				\begin{tabular}{c c c}
					\toprule
					                       \textbf{Augmentation}                         & \textbf{Dataset} & \textbf{$\text{Match}_m$} \\
					                         \textbf{Approach}                           & \textbf{modified (\%)}              & \textbf{Score}            \\ \midrule
					                                None                                 & 0                              & 0.711                     \\ \midrule
					\multirow{3}{*}{\shortstack[c]{Word removal\\($\le$1 per sentence)}} & 20                             & 0.705                     \\
					                                                                     & 50                             & 0.702                     \\
					                                                                     & 100                            & $\mathbf{0.716}$          \\ \midrule
					\multirow{4}{*}{\shortstack[c]{Word removal\\($\ge$1 per sentence)}} & 20                             & 0.711                     \\
					                                                                     & 50                             & 0.704                     \\
					                                                                     & 60                             & 0.712                     \\
					                                                                     & 70                             & 0.705                     \\ \midrule
					                        Upper-casing a word                          & 30                             & 0.687                     \\ \midrule
					              \multirow{2}{*}{Reversing the sentence}                & 10                             & 0.704                     \\
					                                                                     & 100                            & 0.707                     \\ \bottomrule
				\end{tabular}%
			}
			\caption{Data Augmentation Analysis}\label{tab:dataAugmentationAnalysis}
		\end{table}
		
		\subsection{Data Augmentation Analysis}\label{subsec:dataAugmentationAnalysisResults}
			Table \ref{tab:dataAugmentationAnalysis} shows that there is a little score gain by applying data augmentation techniques. For each strategy, we experiment by applying the augmentation approach to different percentages of the total training data and calculated $\text{Match}_m$ score. We observe that word upper-casing strategy results in a significant drop in the score due to model overfitting whereas word removal strategy (maximum 1 word per sentence) on entire training data gives highest $\text{Match}_m$ score of 0.716.

	\section{Conclusion}\label{sec:conclusion}
		Modeling lightweight neural models that can run on low-resource devices can greatly enhance the end-user experience. In this work, we propose a novel, lightweight EmpLite model for text emphasis selection that can run on edge devices such as smartphones for choosing prominent words from short, informal text.  We approach the emphasis selection problem as a sequence labeling task and multiple experiments have shown consistent improvement in the accuracy. Our experimental results show the impact of the attention layer and of using POS as an additional feature in boosting the matching score. Our best performing model achieves an overall matching score of 0.716 with a size of 2.82 MB, proving its effectiveness to run on low-resource edge devices.
		
		Future work includes increasing the vocabulary with commonly used words in English and exploring thin versions of BERT like DistilBERT \cite{sanh2019distilbert}, MobileBERT \cite{sun2020mobilebert}, and TinyBERT \cite{jiao-etal-2020-tinybert} for modeling emphasis.

\bibliography{bibliography}\nocite{*}

\begin{thebibliography}{32}
\expandafter\ifx\csname natexlab\endcsname\relax\def\natexlab#1{#1}\fi

\bibitem[{Abadi et~al.(2016)Abadi, Agarwal, Barham, Brevdo, Chen, Citro,
  Corrado, Davis, Dean, Devin et~al.}]{abadi2016tensorflow}
Mart{\'\i}n Abadi, Ashish Agarwal, Paul Barham, Eugene Brevdo, Zhifeng Chen,
  Craig Citro, Greg~S Corrado, Andy Davis, Jeffrey Dean, Matthieu Devin, et~al.
  2016.
\newblock Tensorflow: Large-scale machine learning on heterogeneous distributed
  systems.
\newblock \emph{arXiv preprint arXiv:1603.04467}.

\bibitem[{Adobe(2016)}]{adobe-spark}
Adobe. 2016.
\newblock {Adobe Spark}.
\newblock \url{https://spark.adobe.com}, accessed 2020-11-28.

\bibitem[{Anand et~al.(2020)Anand, Gupta, Yadav, Mahata, Gosangi, Zhang, and
  Shah}]{anand2020midas}
Sarthak Anand, Pradyumna Gupta, Hemant Yadav, Debanjan Mahata, Rakesh Gosangi,
  Haimin Zhang, and Rajiv~Ratn Shah. 2020.
\newblock Midas at semeval-2020 task 10: Emphasis selection using label
  distribution learning and contextual embeddings.
\newblock \emph{arXiv preprint arXiv:2009.02619}.

\bibitem[{Devlin et~al.(2018)Devlin, Chang, Lee, and
  Toutanova}]{devlin2018bert}
Jacob Devlin, Ming-Wei Chang, Kenton Lee, and Kristina Toutanova. 2018.
\newblock Bert: Pre-training of deep bidirectional transformers for language
  understanding.
\newblock \emph{arXiv preprint arXiv:1810.04805}.

\bibitem[{Hochreiter and Schmidhuber(1997)}]{10.1162/neco.1997.9.8.1735}
Sepp Hochreiter and J\"{u}rgen Schmidhuber. 1997.
\newblock \href {https://doi.org/10.1162/neco.1997.9.8.1735} {Long short-term
  memory}.
\newblock \emph{Neural Comput.}, 9(8):1735–1780.

\bibitem[{Huang et~al.(2020)Huang, Feng, Su, Chen, Wang, Liu, Ouyang, and
  Sun}]{huang2020ernie}
Zhengjie Huang, Shikun Feng, Weiyue Su, Xuyi Chen, Shuohuan Wang, Jiaxiang Liu,
  Xuan Ouyang, and Yu~Sun. 2020.
\newblock Ernie at semeval-2020 task 10: Learning word emphasis selection by
  pre-trained language model.
\newblock \emph{arXiv preprint arXiv:2009.03706}.

\bibitem[{Jiao et~al.(2020)Jiao, Yin, Shang, Jiang, Chen, Li, Wang, and
  Liu}]{jiao-etal-2020-tinybert}
Xiaoqi Jiao, Yichun Yin, Lifeng Shang, Xin Jiang, Xiao Chen, Linlin Li, Fang
  Wang, and Qun Liu. 2020.
\newblock \href {https://doi.org/10.18653/v1/2020.findings-emnlp.372}
  {{T}iny{BERT}: Distilling {BERT} for natural language understanding}.
\newblock In \emph{Findings of the Association for Computational Linguistics:
  EMNLP 2020}, pages 4163--4174, Online. Association for Computational
  Linguistics.

\bibitem[{Kim(2020)}]{kim2020efficient}
Hyeon~Gyu Kim. 2020.
\newblock Efficient keyword extraction from social big data based on cohesion
  scoring.
\newblock \emph{Journal of the Korea Society of Computer and Information},
  25(10):87--94.

\bibitem[{Kingma and Ba(2014)}]{kingma2014adam}
Diederik~P Kingma and Jimmy Ba. 2014.
\newblock Adam: A method for stochastic optimization.
\newblock \emph{arXiv preprint arXiv:1412.6980}.

\bibitem[{Liang et~al.(2017)Liang, Xu, and Zhao}]{liang2017combining}
Dongyun Liang, Weiran Xu, and Yinge Zhao. 2017.
\newblock Combining word-level and character-level representations for relation
  classification of informal text.
\newblock In \emph{Proceedings of the 2nd Workshop on Representation Learning
  for NLP}, pages 43--47.

\bibitem[{Litvak and Last(2008)}]{litvak2008graph}
Marina Litvak and Mark Last. 2008.
\newblock Graph-based keyword extraction for single-document summarization.
\newblock In \emph{Coling 2008: Proceedings of the workshop Multi-source
  Multilingual Information Extraction and Summarization}, pages 17--24.

\bibitem[{Liu et~al.(2019)Liu, Ott, Goyal, Du, Joshi, Chen, Levy, Lewis,
  Zettlemoyer, and Stoyanov}]{liu2019roberta}
Yinhan Liu, Myle Ott, Naman Goyal, Jingfei Du, Mandar Joshi, Danqi Chen, Omer
  Levy, Mike Lewis, Luke Zettlemoyer, and Veselin Stoyanov. 2019.
\newblock Roberta: A robustly optimized bert pretraining approach.
\newblock \emph{arXiv preprint arXiv:1907.11692}.

\bibitem[{Ma and Hovy(2016)}]{ma2016end}
Xuezhe Ma and Eduard Hovy. 2016.
\newblock End-to-end sequence labeling via bi-directional lstm-cnns-crf.
\newblock \emph{arXiv preprint arXiv:1603.01354}.

\bibitem[{Marcus et~al.(1994)Marcus, Kim, Marcinkiewicz, MacIntyre, Bies,
  Ferguson, Katz, and Schasberger}]{10.3115/1075812.1075835}
Mitchell Marcus, Grace Kim, Mary~Ann Marcinkiewicz, Robert MacIntyre, Ann Bies,
  Mark Ferguson, Karen Katz, and Britta Schasberger. 1994.
\newblock \href {https://doi.org/10.3115/1075812.1075835} {The penn treebank:
  Annotating predicate argument structure}.
\newblock In \emph{Proceedings of the Workshop on Human Language Technology},
  HLT '94, page 114–119, USA. Association for Computational Linguistics.

\bibitem[{Matsuo and Ishizuka(2004)}]{matsuo2004keyword}
Yutaka Matsuo and Mitsuru Ishizuka. 2004.
\newblock Keyword extraction from a single document using word co-occurrence
  statistical information.
\newblock \emph{International Journal on Artificial Intelligence Tools},
  13(01):157--169.

\bibitem[{MonkeyLearn(2020)}]{monkeylearn-keyword-extraction}
MonkeyLearn. 2020.
\newblock {Keyword Extraction: A Guide to Finding Keywords in Text}.
\newblock \url{https://monkeylearn.com/keyword-extraction/}, accessed
  2020-07-03.

\bibitem[{NEON(2020)}]{samsung-neon}
NEON. 2020.
\newblock {NEON}.
\newblock \url{https://www.neon.life/}, accessed 2020-06-24.

\bibitem[{Pennington et~al.(2014)Pennington, Socher, and
  Manning}]{pennington-etal-2014-glove}
Jeffrey Pennington, Richard Socher, and Christopher Manning. 2014.
\newblock \href {https://doi.org/10.3115/v1/D14-1162} {{G}lo{V}e: Global
  vectors for word representation}.
\newblock In \emph{Proceedings of the 2014 Conference on Empirical Methods in
  Natural Language Processing ({EMNLP})}, pages 1532--1543, Doha, Qatar.
  Association for Computational Linguistics.

\bibitem[{Peters et~al.(2018)Peters, Neumann, Iyyer, Gardner, Clark, Lee, and
  Zettlemoyer}]{peters2018deep}
Matthew~E Peters, Mark Neumann, Mohit Iyyer, Matt Gardner, Christopher Clark,
  Kenton Lee, and Luke Zettlemoyer. 2018.
\newblock Deep contextualized word representations.
\newblock \emph{arXiv preprint arXiv:1802.05365}.

\bibitem[{Quotes(2020)}]{wisdom-quotes}
Wisdom Quotes. 2020.
\newblock {Wisdom Quotes - Get wiser slowly}.
\newblock \url{https://wisdomquotes.com}, accessed 2020-11-30.

\bibitem[{RiTUAL-UH(2020)}]{SemEval2020-Task10-Emphasis-Selection}
RiTUAL-UH. 2020.
\newblock {SemEval2020 Task10 Emphasis Selection}.
\newblock
  \url{https://github.com/RiTUAL-UH/SemEval2020_Task10_Emphasis_Selection},
  accessed 2020-11-22.

\bibitem[{Rose et~al.(2010)Rose, Engel, Cramer, and Cowley}]{rose2010automatic}
Stuart Rose, Dave Engel, Nick Cramer, and Wendy Cowley. 2010.
\newblock Automatic keyword extraction from individual documents.
\newblock \emph{Text mining: applications and theory}, 1:1--20.

\bibitem[{Sanh et~al.(2019)Sanh, Debut, Chaumond, and
  Wolf}]{sanh2019distilbert}
Victor Sanh, Lysandre Debut, Julien Chaumond, and Thomas Wolf. 2019.
\newblock Distilbert, a distilled version of bert: smaller, faster, cheaper and
  lighter.
\newblock \emph{arXiv preprint arXiv:1910.01108}.

\bibitem[{shallowLearner(2020)}]{semeval2020-task10}
shallowLearner. 2020.
\newblock {SemEval 2020 - Task 10: Emphasis Selection For Written Text in
  Visual Media}.
\newblock \url{https://competitions.codalab.org/competitions/20815}, accessed
  2020-08-15.

\bibitem[{Shirani et~al.(2019)Shirani, Dernoncourt, Asente, Lipka, Kim,
  Echevarria, and Solorio}]{shirani2019learning}
Amirreza Shirani, Franck Dernoncourt, Paul Asente, Nedim Lipka, Seokhwan Kim,
  Jose Echevarria, and Thamar Solorio. 2019.
\newblock Learning emphasis selection for written text in visual media from
  crowd-sourced label distributions.
\newblock In \emph{Proceedings of the 57th Annual Meeting of the Association
  for Computational Linguistics}, pages 1167--1172.

\bibitem[{Singhal et~al.(2020)Singhal, Dhull, Agarwal, and
  Modi}]{singhal2020iitk}
Vipul Singhal, Sahil Dhull, Rishabh Agarwal, and Ashutosh Modi. 2020.
\newblock Iitk at semeval-2020 task 10: Transformers for emphasis selection.
\newblock \emph{arXiv preprint arXiv:2007.10820}.

\bibitem[{Sun et~al.(2020{\natexlab{a}})Sun, Wang, Li, Feng, Tian, Wu, and
  Wang}]{sun2020ernie}
Yu~Sun, Shuohuan Wang, Yu-Kun Li, Shikun Feng, Hao Tian, Hua Wu, and Haifeng
  Wang. 2020{\natexlab{a}}.
\newblock Ernie 2.0: A continual pre-training framework for language
  understanding.
\newblock In \emph{AAAI}, pages 8968--8975.

\bibitem[{Sun et~al.(2020{\natexlab{b}})Sun, Yu, Song, Liu, Yang, and
  Zhou}]{sun2020mobilebert}
Zhiqing Sun, Hongkun Yu, Xiaodan Song, Renjie Liu, Yiming Yang, and Denny Zhou.
  2020{\natexlab{b}}.
\newblock Mobilebert: a compact task-agnostic bert for resource-limited
  devices.
\newblock \emph{arXiv preprint arXiv:2004.02984}.

\bibitem[{Vaswani et~al.(2017)Vaswani, Shazeer, Parmar, Uszkoreit, Jones,
  Gomez, Kaiser, and Polosukhin}]{vaswani2017attention}
Ashish Vaswani, Noam Shazeer, Niki Parmar, Jakob Uszkoreit, Llion Jones,
  Aidan~N Gomez, {\L}ukasz Kaiser, and Illia Polosukhin. 2017.
\newblock Attention is all you need.
\newblock \emph{Advances in neural information processing systems},
  30:5998--6008.

\bibitem[{Yang et~al.(2019)Yang, Dai, Yang, Carbonell, Salakhutdinov, and
  Le}]{yang2019xlnet}
Zhilin Yang, Zihang Dai, Yiming Yang, Jaime Carbonell, Russ~R Salakhutdinov,
  and Quoc~V Le. 2019.
\newblock Xlnet: Generalized autoregressive pretraining for language
  understanding.
\newblock In \emph{Advances in neural information processing systems}, pages
  5753--5763.

\bibitem[{Zhang(2008)}]{zhang2008automatic}
Chengzhi Zhang. 2008.
\newblock Automatic keyword extraction from documents using conditional random
  fields.
\newblock \emph{Journal of Computational Information Systems}, 4(3):1169--1180.

\bibitem[{Zhang et~al.(2006)Zhang, Xu, Tang, and Li}]{zhang2006keyword}
Kuo Zhang, Hui Xu, Jie Tang, and Juanzi Li. 2006.
\newblock Keyword extraction using support vector machine.
\newblock In \emph{international conference on web-age information management},
  pages 85--96. Springer.

\end{thebibliography}
\bibliographystyle{acl_natbib}

\end{document}